# FontNet: Closing the gap to font designer performance in font synthesis

Ammar Ul Hassan
Soongsil University
ammar.instantsoft@gmail.com

Jaeyoung Choi
Soongsil University
choi@ssu.ac.kr

## Abstract

*Font synthesis has been a very active topic in recent years because manual font design requires domain expertise and is a labor-intensive and time-consuming job. While remarkably successful, existing methods for font synthesis have major shortcomings; they require finetuning for unobserved font style with large reference images, the recent few-shot font synthesis methods are either designed for specific language systems or they operate on low-resolution images which limits their use. In this paper, we tackle this font synthesis problem by learning the font style in the embedding space. To this end, we propose a model, called FontNet, that simultaneously learns to separate font styles in the embedding space where distances directly correspond to a measure of font similarity, and translates input images into the given observed or unobserved font style. Additionally, we design the network architecture and training procedure that can be adopted for any language system and can produce high-resolution font images. Thanks to this approach, our proposed method outperforms the existing state-of-the-art font generation methods on both qualitative and quantitative experiments.*

## 1. Introduction

Font design is critical in helping readers in interpreting information contained in the text. Manual font design, on the other hand, requires domain expertise and is a labor-intensive and time-consuming task. Font design becomes more complicated and expensive as the number of characters in a language increase, such as Chinese (over 50,000 glyphs) or Korean (11, 172 glyphs).

With the advancements in the generated image quality of generative adversarial network (GAN) [1], recent font generation methods based on GANs have shown rapid improvement [2, 3, 4, 5, 6, 7, 8, 9, 10, 11, 12]. Yet each of these methods has some limitations, including the need to finetune the pretrained model on a large number of unobserved font images to generate a new font set [2, 3, 6]. Recent methods solve this prior problem by proposing few-shot font generation (FFG) frameworks [5, 8-12] that require very few unobserved font images during inference.

Table 1: Comparison of existing font synthesis methods with our model. We list some of the limitations of existing font generation methods

| Font generation Methods | Require Finetuning? | Regularize Style Encoder? | Language specific? | Low image resolution? |
|---|---|---|---|---|
| Zi2zi [2] | yes | no | no | no |
| DC-Font [3] | yes | no | no | no |
| SK-Font [4] | yes | no | no | no |
| MC-GAN [5] | no | no | yes | yes |
| AGIS-Net [6] | yes | no | no | yes |
| DM-Font [8] | no | no | yes | no |
| LF-Font [9] | no | no | yes | no |
| MX-Font [10] | no | yes | yes | no |
| DG-Font [11] | no | no | no | yes |
| FtransGAN [12] | no | no | yes | yes |
| FontNet (Ours) | no | yes | no | no |

Despite their efforts, the architectures of these methods are either designed for specific language systems [5, 8-10] or they operate on low-resolution images [5, 11, 12] that hinders their practical usage. Additionally, we observed that most of these methods (except MX-Font) do not regularize the style encoders directly. For learning font style, these methods mostly rely on multi-task discriminator [13] that is trained to classify each font style in an adversarial manner. The multi-task discriminator utilizes font style labels to provide useful gradients to the generator for font style transfer. FtransGAN [12] utilizes images rather than labels to learn style using PatchGAN discriminator [14]. **Table 1** presents the comparisons of these state-of-the-art (SOTA) methods.

As a result of the aforementioned observations, we propose an effective FFG method, named FontNet, that does not require the constraints listed in Table 1. We design a simple network architecture that can be adopted for any language system. To overcome the issue of generating high-resolution font images, we utilize the SOTA StyleGAN [15] model modified for this FFG task. We introduce a novel font separator network inspired from the metric learning literature that learns the font style similarity in the embedding space. Rather than classifying the font style during training, our separator network learns to group characters of the same font style closer to each other and apart from other font styles in the embedding space. This strategy enables us to generate an unobserved font set with only a few samples (we used 8 samples in our experiments). We validate our proposed FontNet quantitatively and qualitatively on Korean hangul characters. We demonstrate in our experiments that our method outperforms SOTA FFG baselines both qualitatively and quantitatively.

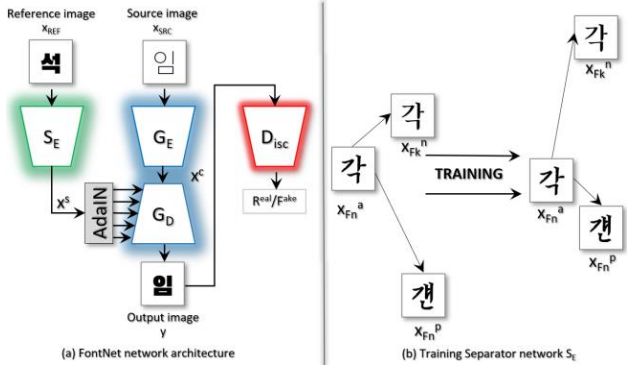

Figure 1: Architecture of FontNet is presented in (a). The style features $x^s$ extracted from separator network $S_E$ are injected via AdaIN into the generator G which decodes the content features $x^c$ and generates output image y depicting the style of reference image $X_{REF}$ and content of source image $x_{SRC}$. In (b) we visualize the loss function used for training the separator network. Note that for only demonstration we show the images ($x_{Fn}^a$, $x_{Fn}^p$, $x_{Fk}^n$) in (b) for actual training we use the style embeddings more details in section 2.2.

## 2. FontNet

### 2.1. Architecture Overview

The network architecture of FontNet is presented in Figure 1(a). In our framework, the separator network ($S_E$ in Figure 1(a)) plays a key role as a font style encoder by learning the font similarity in the font style embedding space. It guides the font synthesis by feeding the font style features $x^s$ of a reference image $x_{REF}$ to the generator G. The generator G synthesizes an output image y with the style from the reference image $x_{REF}$ and the content of the source image $x_{SRC}$. The discriminator $D_{isc}$ is also employed for better generation quality.

### 2.2. Learning to encode font style features

Given a reference image representing the style of the target font, the style encoder needs to understand and extract the style features of the given image. Existing methods [2-6, 8, 9, 11, 12] only use the feedback from the style conditional discriminators trained over a set of known font styles to guide the generator to synthesize the target font style.

Different from these approaches that learn style using style classification (in an adversarial manner), we train our separator network $S_E$ to learn the font style similarity in the embedding space. To this end, we employ triplet loss [16] in the separator network $S_E$ that extracts style embedding f(x), from a given font image x into an embedding space $R^d$, such that the square distance between all font images belonging to the same font style is small, whereas the squared distance between a pair of font images belonging to different font style is large. Here, in order to learn the font style embedding, we train the separator network $S_E$ such that a font image $x_{Fn}^a$ (anchor) belonging to a specific font style $F_n$ is closer to all other font images $x_{Fn}^p$ (positive) of the same font style $F_n$ than it is to any font image $x_{Fk}^n$ (negative) of a different font style $F_k$. Our separator network $S_E$ triplet loss is given by:

$$\mathcal{L}_{enc_{style}} = \|f(x_{Fn}^a) - f(x_{Fn}^p)\|_2^2 \\ - \|f(x_{Fn}^a) - f(x_{Fk}^n)\|_2^2 + \alpha, \quad (1)$$

where f($x_{Fn}^a$), f($x_{Fn}^p$), and f($x_{Fk}^n$) are the embeddings of anchor, positive, and negative images, respectively, and $\alpha$ is a margin between positive and negative pairs. For injecting the style features $x^s$ of a given image into the generator G, we extract features from the intermediate layer of the separator network $S_E$ which is later used to guide the generator G for font synthesis. During training, we choose the triplets such that the distance between the anchor and negative is smaller than the distance between the anchor and positive. For implementing this hard triplet mining, we choose the anchor and positive such that they belong to the same font style $F_n$ but with different character content, whereas we chose the negative sample such that it has the same character content with the anchor image but different font style $F_k$. The loss in equation 1 is visualized in Figure 1(b). This loss helps in enhancing the representation power of the style encoder ($S_E$) and affects the generator G for efficient font style transfer.

### 2.3. Learn to synthesize font images

For generating high-resolution font images, we build our generator architecture based on StyleGAN [15]. However, StyleGAN is an unconditional model whereas font synthesis requires desired content and style control. To overcome this issue, we modify the StyleGAN model by adding an encoder ($G_E$ in Figure 1(a)). We replace the default constant input of StyleGAN model with the content features $x^c$ extracted from the content encoder $G_E$. We also remove the mapping network and G regularization from the default StyleGAN generator. We train our generator G as an auto-encoder (encoder $G_E$ and decoder $G_D$) with a discriminator $D_{isc}$ which is same as StyleGAN. The encoder $G_E$ takes a source image $x_{SRC}$ representing the content of the output image and extracts the content features $x^c$. These content features $x^c$ are then fed into the decoder $G_D$. Additionally, the style features $x^s$ extracted from the separator network $S_E$ are injected using AdaIN in all layers of the decoder $G_D$ to generate the output image y representing the style from the reference font image $x_{REF}$ and content from the source image $x_{SRC}$. To this end, we adopt the following losses for training the generator G.

For better visual quality of the generated font images, we train the generator in an adversarial manner with a discriminator. The discriminator adopts multi-task discriminator $D_y$ [13] for conditioning the character content (classifying character content) and style (classifying character style) in an adversarial manner.

$$\mathcal{L}_{adv} = \mathbb{E}_{x,y}[\log D_y(x^{gt})] + \mathbb{E}_{x,y}\left[\log\left(1 - D_y(G(x^c, x^s))\right)\right]. \quad (2)$$

In order to prevent a degenerate case where the generator G ignores the style features $x^s$ from the separator network $S_E$ and synthesizes an image of random font style, we introduce a style triplet loss:

$$\mathcal{L}_{g_{style}} = \|f(y_{Fn}^a) - f(x_{Fn}^p)\|_2^2 - \|f(y_{Fn}^a) - f(x_{Fk}^n)\|_2^2 + \alpha. \quad (3)$$

Here the triplet loss is employed for forcing the generator G to adopt the style features $x^s$ extracted from the separator network $S_E$. The major difference between the equation (1) and equation (3) is that the anchor image used in equation (3) is the generated image $y_{Fn}^a$ while we use a real image $x_{Fn}^a$ in equation (1). This loss guides the generated image y to have a style similar to the reference font image $x_{REF}$ and dissimilar to the other samples (negative).

We also employ pixel level supervision by adding pixel by pixel difference between generated image y and ground truth image $x^{gt}$. The loss is given by.

$$\mathcal{L}_{g_{L1}} = \mathbb{E}_{x,y}[\|x^{gt} - y\|_1] \quad (4)$$

The final objective function to train the generator, the discriminator, and the separator network is given by:

$$\min_{G,S_E} \max_{D} \mathcal{L}_{adv} + \lambda \mathcal{L}_{g_{L1}} + \lambda \mathcal{L}_{g_{style}} + \lambda \mathcal{L}_{enc_{style}} \quad (5)$$

## 3. Experiments

In this section, we validate our proposed FontNet on Korean hangul characters and compare it with state-of-the-art FFG methods.

### 3.1. Datasets

We build our dataset by downloading 90 Korean fonts from Naver[1]. We use 75% fonts for training and remaining for testing. Additionally, we use the most common 2,350 Korean characters [17] and randomly split 2,000 for training (seen characters) and remaining 350 for testing (unseen characters). We separately evaluate the models on seen and unseen characters to measure the generalizability.

### 3.2. Baselines and Evaluation metrics

For comparing our model with state-of-the-art methods listed in Table 1, we exclude the methods which require extra finetuning steps [2-4, 6] or operate on low-resolution images [5, 11, 12]. We choose recently proposed MX-Font [10], an improved method of its prior variants [8, 9] and FUNIT [13], which is modified for this FFG task. We use the official implementations[2] of these baselines provided by the MX-Font authors.

We evaluate the models on pixel-level, perceptual-level, and content-style classification metrics. For pixel-level evaluation we use the structural similarity (SSIM) metric. For perceptual and content-style classification evaluation, we train character and style classifiers using character and style supervision. We then use the mean FID (mFID) and top-1 accuracy (Acc) from the classifiers as the perceptual and content-style classification, respectively.

### 3.3. Quantitative evaluation

Table 2 summarizes the quantitative results. FontNet outperforms baselines on the majority of evaluation metrics, particularly style-aware metrics, in the experiments. On pixel-level evaluation (SSIM), the baselines show slightly worse performance on unseen characters than the seen characters MX-Font (0.693 → 0.677) and FUNIT (0.664 → 0.658) in Table 2. MX-Font performs better on seen and unseen characters for content classification metric than ours however, for the content perceptual score (mFID) our FontNet performs better on seen and unseen characters. Additionally, our method clearly outperforms the baselines in both style-aware metrics. Our model achieves an accuracy of 76.5 (seen characters) and 74.0 (unseen characters) for style classification, compared to MX-Font which has 70.2 (seen characters) and 67.7 (unseen characters). Similarly, FontNet performs better in style perceptual score (mFID) than the baselines.

Table 2: Quantitative evaluation on the dataset described in Section 3.1. All methods are evaluated on both seen and unseen characters. The best result is denoted by a bold number.

| | SSIM ↑ | mFID (C) ↓ | Acc (C) ↑ | mFID (S) ↓ | Acc(S) ↑ |
|---|---|---|---|---|---|
| **seen characters during training** | | | | | |
| MX-Font | 0.693 | 9.3 | **98.5** | 27.6 | 70.2 |
| FUNIT | 0.664 | 11.6 | 96.1 | 49.2 | 28.2 |
| FontNet (ours) | **0.742** | **8.9** | 97.2 | **26.0** | **76.5** |
| **unseen characters during training** | | | | | |
| MX-Font | 0.677 | 14.2 | **95.6** | 54.7 | 67.7 |
| FUNIT | 0.658 | 21.6 | 94.8 | 86.4 | 29.0 |
| FontNet (ours) | **0.746** | **12.3** | 94.6 | **48.4** | **74.0** |

---

[1] The public fonts are collected from https://hangeul.naver.com/font

[2] We use the baselines official implementations from https://github.com/clovaai/fewshot-font-generation

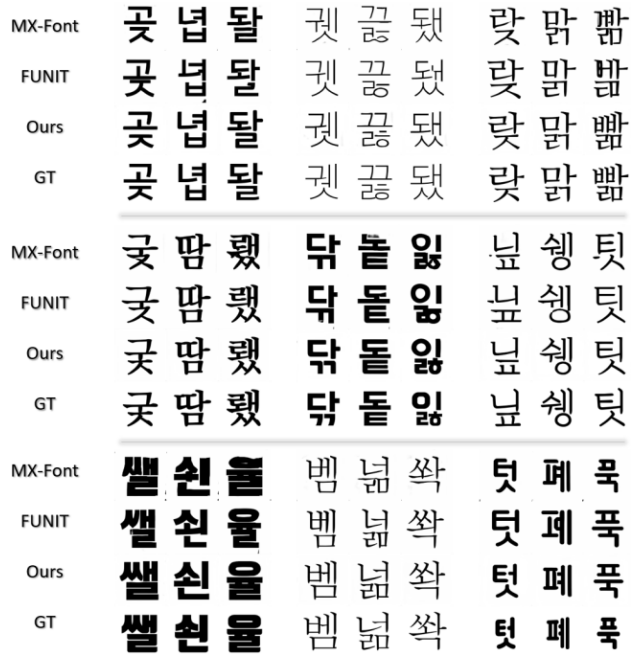

Figure 2: Qualitative results of our proposed method and FFG baselines are presented. GT represents the ground truth font image.

Table 3: Ablation study on the proposed two key ideas in FontNet. w/o our $S_E$ represents a setting of the proposed method with no separator network $S_E$ and the proposed triplet loss. We replace $S_E$ with style encoder from FUNIT [13] as a part of Generator G. w/o our $G_D$ represents a setting of the proposed method without our decoder $D_E$ based on StyleGAN [15]. We replace the decoder with the decoder proposed in FUNIT [13]. The default proposed setting renders better performances (bold numbers) on all evaluation metrics.

|  | SSIM ↑ | mFID (C) ↓ | Acc (C) ↑ | mFID (S) ↓ | Acc(S) ↑ |
|---|---|---|---|---|---|
| *seen characters during training* | | | | | |
| w/o our $S_E$ | 0.693 | 14.9 | 93.57 | 73.5 | 50.52 |
| w/o our $G_D$ | 0.729 | 12.72 | 90.87 | 48.8 | 63.29 |
| proposed | **0.742** | **8.9** | **97.2** | **26.0** | **76.5** |
| *unseen characters during training* | | | | | |
| w/o our $S_E$ | 0.718 | 19.82 | 90.97 | 88.2 | 46.70 |
| w/o our $G_D$ | 0.743 | 14.3 | 90.68 | 57.6 | 62.17 |
| proposed | **0.746** | **12.3** | **94.6** | **48.4** | **74.0** |

### 3.4. Qualitative comparison

The qualitative results are presented in Figure 2. All methods were evaluated on various challenging fonts based on thickness, thinness, and serifs. For this experiment we use a fixed source font that is transformed into target font styles (shown in ground truth (GT) rows) by all methods. As demonstrated in the Figure 2, MX-Font and FUNIT accurately generate the global font styles however the generated content is often broken or noisy. MX-Font completely fails to generate thick fonts (row 9, column 1, 2, 3). For cursive fonts with strokes both baselines fail to generate the local character component content and styles (row 1, 2, column 7). Compared to the baseline methods, our FontNet successfully generates thick fonts, cursive fonts with strokes and other fonts in terms of global and local font content and styles close to the ground truth fonts.

### 3.5. Ablation study

FontNet mainly introduces two key ideas: First, training the separator network with triplet regularization for grouping each font style separately in the style embedding space. Second, the modified StyleGAN generator for generating high quality images. In this section, we perform an ablation study where we remove both key components individually and replace them with traditional approaches, i.e., first replacing separator network $S_E$ with style encoder (style encoder based on FUNIT [13]) which is jointly trained with our generator G. Secondly, replacing our decoder $G_E$ with a traditional decoder (based on FUNIT [13]).

Table 3 presents the results of this ablation study analyzing the impact of our key ideas. We can see that removing the separator network $S_E$ degrades the performance on all evaluation metrics, especially, the style-aware metrics. The style accuracy drops significantly for both seen and unseen font characters. Without our proposed decoder $G_D$ which is based on StyleGAN we observe that the content classification accuracy drops for both seen and unseen characters. For all evaluation metrics, the proposed method performs significantly better. This experiment indicates that our proposed key components effectively help in improving the font synthesis performance of our model in terms of all metrics.

### 4. Conclusion

In this paper, we discussed various shortcomings of existing font generation methods. We proposed a generalized method, FontNet, for font generation task that can be used for any writing system. By exploiting representation learning, FontNet learns the style features of few-shot reference font images. We also employ a modified StyleGAN generator architecture for generating high-quality font images. The experimental results demonstrate that FontNet performs better than the existing font generation methods in terms of qualitative and quantitative experiments.